\documentclass[sigconf]{acmart}
\usepackage{algorithm, algpseudocode}

\begin{document}

% \author{Ben Trovato}
% \authornote{Both authors contributed equally to this research.}
% \email{trovato@corporation.com}
% \orcid{1234-5678-9012}
% \author{G.K.M. Tobin}
% \authornotemark[1]
% \email{webmaster@marysville-ohio.com}
% \affiliation{%
%   \institution{Institute for Clarity in Documentation}
%   \streetaddress{P.O. Box 1212}
%   \city{Dublin}
%   \state{Ohio}
%   \country{USA}
%   \postcode{43017-6221}
% }
%%
%% The "title" command has an optional parameter,
%% allowing the author to define a "short title" to be used in page headers.
\title{The CSIRO Crown-of-Thorn Starfish Detection Dataset}

\settopmatter{printacmref=false} % Removes citation information below abstract
\renewcommand\footnotetextcopyrightpermission[1]{} % removes footnote with conference information in first column
\pagestyle{plain} % removes running headers
\setcopyright{none}
%%
%% The "author" command and its associated commands are used to define
%% the authors and their affiliations.
%% Of note is the shared affiliation of the first two authors, and the
%% "authornote" and "authornotemark" commands
%% used to denote shared contribution to the research.

% jiajun liu
% brano kusy
% ross merchant
% Brendan Do
% Torsten Merz
% Joey Crosswell
% lachlan tychsen-smith
% David Ahmedt Aristizabal david.ahmedtaristizabal@data61.csiro.au
% Andy steven
% geoff carlin
% russ babcock
% peyman moghadam

% \author{Jiajun Liu$^\dag$}
% \author{Brano Kusy$^\dag$}
% \author{Ross Merchant$^{\circ\dag}$}
% \author{Brendan Do}
% \author{Torsten Merz}
% \author{Joey Crosswell}
% \author{Lachlan Tychsen-Smith}
% \author{David Ahmedt Aristizabal}
% \author{Andy Steven}
% \author{Geoff Cralin}
% \author{Russ Babcock}
% \author{Peyman Moghadam}

% \author[1]{Jiajun Liu}
% \author[1]{Brano Kusy}
% \author[1]{Ross Merchant}
% \author[1]{Brendan Do}
% \author[1]{Torsten Merz}
% \author[1]{Joey Crosswell}
% \author[1]{Lachlan Tychsen-Smith}
% \author[1]{David Ahmedt Aristizabal}
% \author[1]{Andy Steven}
% \author[1]{Geoff Cralin}
% \author[1]{Russ Babcock}
% \author[1]{Peyman Moghadam}
% \author[1]{Megha Malpani}

% \affil[1]{Affil1}
% \affil[2]{Affil2}
% \affiliation{~} % to force dual affiliation

\author{Jiajun Liu$^\dag$\quad Brano Kusy$^\dag$\quad Ross Marchant$^{\circ\dag}$\quad  Brendan Do$^\dag$\quad  Torsten Merz$^\dag$\quad  Joey Crosswell$^\star$\newline Andy Steven$^\star$\quad Nic Heaney$^\dag$\quad
Karl von Richter$^\dag$\quad
Lachlan Tychsen-Smith$^\dag$\newline David Ahmedt-Aristizabal$^\dag$\quad Mohammad Ali Armin$^\dag$\quad  Geoffrey Carlin$^\star$\quad  Russ Babcock$^\star$\quad \newline Peyman Moghadam$^\dag$\quad Daniel Smith$\dag$\quad Tim Davis$^\diamond$\quad Kemal El Moujahid$^\diamond$\quad \newline Martin Wicke$^\diamond$\quad Megha Malpani$^\diamond$}

\authornote{Affiliations: $^\dag$CSIRO's Data61, $^\star$CSIRO Oceans \& Atmosphere, $^\circ$Queensland University of Technology, $^\diamond$Google \newline Emails: \{jiajun.liu,\ brano.kusy,\ ross.marchant,\ brendan.do,\ torsten.merz,\ joey.crosswell,\ andy.steven,\ nic.heaney,\ karl.vonrichter,\ laclan.tychsen-smith,\  david.ahmedtaristizabal,\ ali.armin,\ geoffrey.carlin,\ russ.babcock,\ peyman.moghadam,\ daniel.v.smith\}@csiro.au,\ \{timdavis,\ kelmoujahid,\ wicke,\ mmalpani\}@google.com}

\begin{abstract}
  Crown-of-Thorn Starfish (COTS) outbreaks are a major cause of coral loss on the Great Barrier Reef (GBR) and substantial surveillance and control programs are underway in an attempt to manage COTS populations to ecologically sustainable levels. We release a large-scale, annotated underwater image dataset from a COTS outbreak area on the GBR, to encourage research on Machine Learning and AI-driven technologies to improve the detection, monitoring, and management of COTS populations at reef scale. The dataset is released and hosted in a Kaggle competition that challenges the international Machine Learning community with the task of COTS detection from these underwater images \footnote{\url{https://www.kaggle.com/c/tensorflow-great-barrier-reef}}.
\end{abstract}

%%
%% The code below is generated by the tool at http://dl.acm.org/ccs.cfm.
%% Please copy and paste the code instead of the example below.
%%

%%
%% Keywords. The author(s) should pick words that accurately describe
%% the work being presented. Separate the keywords with commas.
%\keywords{datasets, neural networks, gaze detection, text tagging}

%% A "teaser" image appears between the author and affiliation
%% information and the body of the document, and typically spans the
%% page.

\begin{teaserfigure}
  \includegraphics[width=\textwidth]{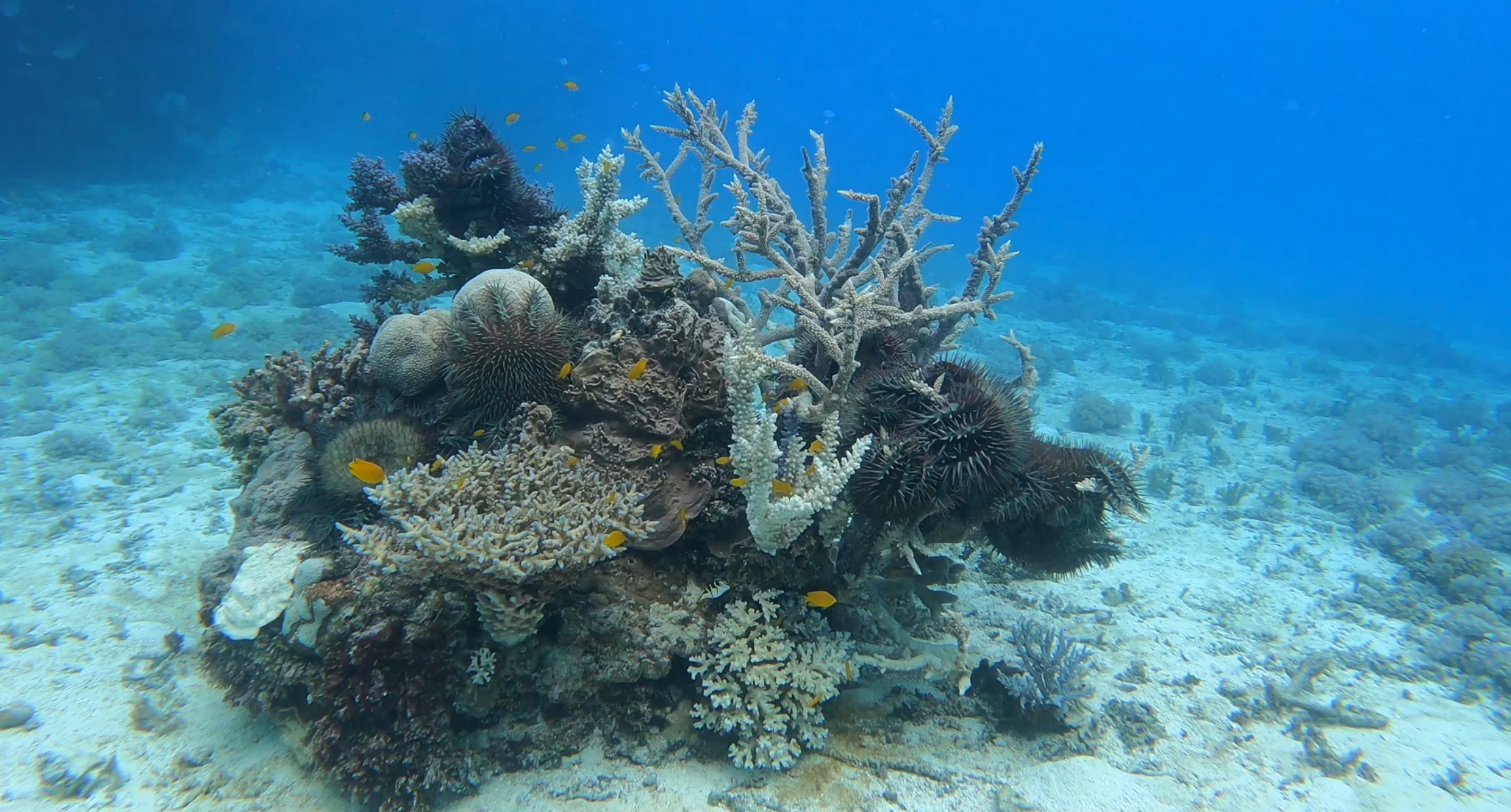}
  \caption{A COTS outbreak at the Swain Reefs, in North Queensland, Australia, October 2021. }
  %\Description{A COTS outbreak at the Swain Reefs, October 2021.}
  \label{fig:teaser}
\end{teaserfigure}
%%
%% This command processes the author and affiliation and title
%% information and builds the first part of the formatted document.
\maketitle

\section{Introduction}

Australia's Great Barrier Reef (GBR) is a national icon and a World Heritage Site, where tiny corals build continental-scale underwater structures inhabited by thousands of other marine creatures. While it is no secret that the GBR is under threat, it may not occur to everyone that a species of starfish, the Crown-of-Thorns Starfish (COTS), is one of the few main factors responsible for coral loss on the GBR.

\begin{figure*}[ht]
  \includegraphics[width=\textwidth]{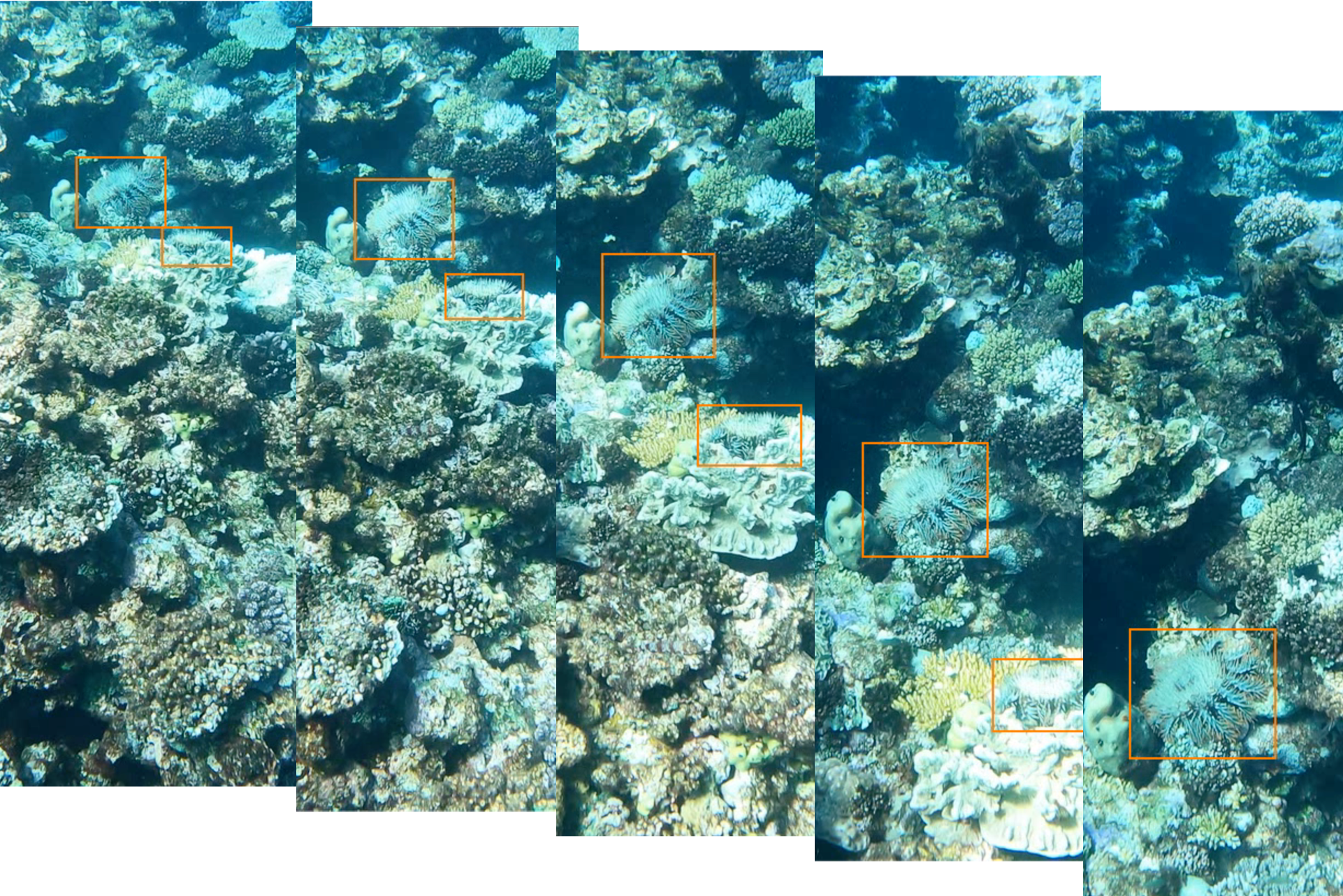}
  \caption{Bounding boxes of the detected COTS are shown in the image sequence.}
  %\Description{A COTS outbreak at the Great Barrier Reef, October 2021.}
  \label{fig:det}
\end{figure*}

Hat-sized and covered with poisonous spines, COTS are a common reef species which feed on corals and have been an integral part of coral reef ecosystems for millennia. However, their populations can explode to thousands of starfish on individual reefs, eating up most of the coral and leaving behind a white-scarred reef that will take years to recover. For centuries the reef was resilient enough to recover from such a natural life cycle, and affected coral was given time to eventually grow back. This is no longer the case as a combination of stress-factors now threaten health of the GBR. The World Wide Fund for Nature (WWF) lists five key threats to the GBR\footnote{\url{https://www.wwf.org.au/what-we-do/oceans/great-barrier-reef\#gs.8nv5fh}} and the COTS outbreaks and their links to increased farm pollution are at the top of the list, second only to the massive bleaching events that impacted the reef in recent history.

Australian agencies are running a major monitoring and management program in an effort to track and control COTS populations to ecologically sustainable levels. One of the key methods used for COTS and coral cover surveillance is the Manta Tow method, in which a snorkel-diver is towed behind a boat to perform visual assessment of the underwater habitat. The boat stops after every transect (typically every 200 meters) to allow variables observed during the tow to be recorded on a datasheet. Information from such surveys is used to identify early outbreaks of COTS or characterize existing outbreaks, and provides key information for the decision support systems to better target deployment of COTS control teams. While effective in general, the survey method has a few limitations related to its operational scalability, data resolution, reliability and traceability.

One way of improving the survey method is to deploy underwater imaging devices to collect data from the reefs and employ deep learning algorithms to analyze the imagery, for example, to find COTS and characterize coral cover. AI-driven environmental surveys could significantly improve the efficiency and scale at which marine scientists and reef managers survey for COTS, offering great benefits to the COTS control program.

The dataset presented in this paper aims to make a bold statement about the capabilities of Machine Learning and AI technologies applied to  broadscale surveillance of underwater habitats. In partnership with  international community, we want to demonstrate a step-change in delivering accurate, timely, and reliable assessment of reef-scale ecosystems. We are releasing a large annotated underwater imagery dataset collected on the GBR, collected in collaboration with COTS Control teams in a COTS outbreak area. We have worked with domain experts and marine scientists to validate the dataset annotations and are sharing the data with international community through a Kaggle competition.

\section{Data Collection Method}
We have collected the data with GoPro Hero9 cameras that were adapted for use in the Manta Tow method. Specifically, the camera was attached to the bottom of the manta tow board that the snorkeler-diver holds onto during the surveys. The camera provides an oblique field of view of the reef below the diver and the distance to the reef dynamically changes as the diver explores the reef. The distance is typically several meters from the bottom, but can be as close as a few tens of centimeters or as far as 10 meters or more. The survey boat moves at a speed up to 5 knots and it pulls the diver for a duration of two minutes which equates to transect of approximately 200 meters. The boat then stops to let the diver record data observed during the transect on a sheet of paper. We set the GoPro cameras to record videos continuously at 24 frames per second at 3840x2160 resolution and manually removed the periods of no activity between transects.

The dataset then underwent an AI-assisted annotation and quality assurance process, in which expert annotators, with the help of pre-trained COTS detection models, identified all COTS in the images. Each COTS detection in an image was marked with a rectangular box using our annotation software. %The dataset was collected at various locations over several days in the Swain Reefs region of the GBR in October 2021. Hence there are variations in the location and weather, and subsequently lighting and coral bed conditions.
The dataset was collected over a single day in October 2021 at a reef in the Swain Reefs region of the GBR. There are variations in lighting, visibility, coral habitat, depth, distance from bottom and viewpoint.

\section{Dataset Characterization}
The released dataset consists of sequences of underwater images collected at five different areas on a reef in the GBR. It contains more than 35k images, with hundreds of individual COTS visible. The dataset is split between training and testing data and for the purposes of the competition, we are releasing the training dataset consisting of images annotated with bounding boxes around each COTS (e.g. Fig. \ref{fig:det}). 

There are a few key factors that differentiate this dataset with conventional object detection datasets:

\begin{itemize}
    \item There is only one class, COTS, in this object detection dataset.
    \item The dataset naturally exhibits sequence-based annotations as multiple images are taken of the same COTS as the boat moves past it. Note that the evaluation is still individual detection-based, similar to the COCO challenge.
    \item There could be multiple COTS in the same image and the COTS can overlap each other. COTS are cryptic animals and like to hide, thus only part of the animal might be visible in some images.
    \item The key objective in surveys conducted by the monitoring and control program is to find all visible COTS along the defined transect paths, hence recall is prioritized. The evaluation is therefore based on the average F2-score (see next section for details).
\end{itemize}

\section{Evaluation Protocol}
For a single image, the IoU of a proposed set of object pixels and a set of true object pixels is calculated as:
$IoU(a,b)=\frac{a\cap b}{a\cup b}$.

The procedure that counts True Positives (TP), False Positives (FP), and False Negatives (FN) is defined as:

\begin{algorithmic}[1]
\Procedure{JudgeDetections}{Groundtruth $\mathbb{B}=\{b_i\}$, submission $\mathbb{S}=\{s_i\}$, IoU threshold $IoU$}
\For{each submitted box $s_j \in \mathbb{S}$ in descending order of confidence score}
  \State Find the unmatched groundtruth box $b_i \in \mathbb{B}$ that has the highest IoU $iou$ with $s_j$
  \If{$iou > IoU$}
    \State add $s_j$ to TP
  \Else
    \State add $s_j$ to FP
  \EndIf
    \State Mark groundtruth box $b_i$ as matched
\EndFor
\EndProcedure
\State Add all unmatched $b_i \in \mathbb{B}$ to FN
\end{algorithmic}

Given an IoU threshold $IoU$, the F$_\beta$-score \cite{article} is calculated based on the number of true positives $TP$, false negatives $FN$, and false positives $FP$,
\begin{equation}
F^{IoU}_\beta(\mathbb{S}) = \frac{ (1+\beta^2)\cdot TP(\mathbb{S})} { (1+\beta^2)\cdot TP(\mathbb{S}) + \beta^2 \cdot FN(\mathbb{S}) + FP(\mathbb{S}) }.
\end{equation}
We use $\beta = 2$ for the competition (F2-score).

The F2-score is calculated for IoU thresholds ranging from 0.3 to 0.8 at 0.05 increments, and we then use the average as the metric for the main leaderboard.

\section*{Acknowledgements}
We thank Dan Godoy and the Blue Planet Marine COTS Control Team and research vessel crew, who helped plan voyage logistics and collected GoPro video alongside their manta tow surveys. We also thank Dave Williamson at GBRMPA for providing data and planning support. 

We thank Google for supporting the project in various ways and thank the Google/Kaggle team that includes Scott Riddle, Danial Formosa, Taehee Jeong, Glenn Cameron, Addison Howard, Will Cukierski, Sohier Dane, Ryan Holbrook, and more.

%%
%% The next two lines define the bibliography style to be used, and
%% the bibliography file.
\bibliographystyle{ACM-Reference-Format}
\bibliography{sample-base}

%%
%% If your work has an appendix, this is the place to put it.

\end{document}